\documentclass{ifacconf}

\usepackage{graphicx}
\usepackage{natbib}
\usepackage{float}

\usepackage{comment}
\usepackage{amsmath, amssymb, amsfonts}
\usepackage{algorithm}
\usepackage{algpseudocode}
\usepackage{textcomp}
\usepackage{stfloats}
\usepackage{xcolor}
\usepackage{url}
\usepackage{caption}
\usepackage{subcaption}

\newcommand{\magenta}[1]{{\color{magenta}{#1}}}

\begin{document}
\begin{frontmatter}

%%%%%%%%%%%%%%%%%%%%%%%%%%%%%%%%%%%%%%%%%%%%%%%%%%%%%%%%%%%%%%%%%%%%%%%%%%%%%%%%
%%%%%%%%%%%%%%%%%%%%%%%%%%%%%%%%%%%% Title %%%%%%%%%%%%%%%%%%%%%%%%%%%%%%%%%%%%%

\title{Real-time Point Cloud Data Transmission via L4S for 5G-Edge-Assisted Robotics}

%%%%%%%%%%%%%%%%%%%%%%%%%%%%%%%%%%%%%%%%%%%%%%%%%%%%%%%%%%%%%%%%%%%%%%%%%%%%%%%%
%%%%%%%%%%%%%%%%%%%%%%%%%%%%%%%%%%% Authors %%%%%%%%%%%%%%%%%%%%%%%%%%%%%%%%%%%%

\author[First]{Gerasimos Damigos}
\author[Second]{Achilleas Santi Seisa}
\author[Second]{Nikolaos Stathoulopoulos} 
\author[First]{Sara Sandberg} 
\author[Second]{George Nikolakopoulos}

\address[First]{Ericsson Research, Lule\aa, Sweden (\small gerasimos.damigos@ericsson.com)}
\address[Second]{Robotics and AI Group, 
% Department of Computer, Electrical and Space Engineering, 
Lule\aa\, University of Technology, Sweden}

% \thanks[footnoteinfo]{This project has received funding from the \textcolor{red}{project?}.}

\maketitle

%%%%%%%%%%%%%%%%%%%%%%%%%%%%%%%%%%%%%%%%%%%%%%%%%%%%%%%%%%%%%%%%%%%%%%%%%%%%%%%%
%%%%%%%%%%%%%%%%%%%%%%%%%%%%%%%%%%% Abstract %%%%%%%%%%%%%%%%%%%%%%%%%%%%%%%%%%%

\begin{abstract}
This article presents a novel framework for real-time Light Detection and Ranging (LiDAR) data transmission that leverages rate-adaptive technologies and point cloud encoding methods to ensure low-latency, and low-loss data streaming. The proposed framework is intended for, but not limited to, robotic applications that require real-time data transmission over the internet for offloaded processing. Specifically, the Low Latency, Low Loss, Scalable Throughput (L4S)-enabled \texttt{SCReAM v2} transmission framework is extended to incorporate the \texttt{Draco} geometry compression algorithm, enabling dynamic compression of high-bitrate 3D LiDAR data according to the sensed channel capacity and network load. The low-latency 3D LiDAR streaming system is designed to maintain minimal end-to-end delay while constraining encoding errors to meet the accuracy requirements of robotic applications. We demonstrate the effectiveness of the proposed method through real-world experiments conducted over a public 5G network across multi-kilometer urban environments. The low-latency and low-loss requirements are preserved, while real-time offloading and evaluation of 3D SLAM algorithms are used to validate the framework’s performance in practical use cases.
\end{abstract}

\end{frontmatter}

%%%%%%%%%%%%%%%%%%%%%%%%%%%%%%%%%%%%%%%%%%%%%%%%%%%%%%%%%%%%%%%%%%%%%%%%%%%%%%%%
%%%%%%%%%%%%%%%%%%%%%%%%%%%%%%%%% Introduction %%%%%%%%%%%%%%%%%%%%%%%%%%%%%%%%%
\section{Introduction} \label{intro}
%%%%%%%%%%%%%%%%%%%%%%%%%%%%%%%%%%%%%%%%%%%%%%%%%%%%%%%%%%%%%%%%%%%%%%%%%%%%%%%%

%% WHY 
Modern robotic platforms increasingly rely on high throughput sensors, such as 3D LiDARs, RGB-D cameras, and radars, to perceive and reason about their surroundings. However, running advanced perception and decision-making algorithms on these data streams in real time, is a huge challenge due to the limited onboard computational capability of resource-constrained robotic platforms as highlighted by~\cite{huang2022edge}. Power consumption, payload capacity, and thermal constraints restrict the utilization of high performance computing hardware.
To reduce such limitations, edge-assisted robotics is a forthcoming solution (\cite{seisa2022med}). By offloading compute intensive operations to nearby edge servers, robots can leverage low-latency access to scalable computational resources. This does not only enable the execution of complex algorithms in real time without overloading the onboard processor, but also allows certain tasks to be completed significantly faster due to parallelization and hardware acceleration, even when the transmission latency is taken into account.

%% WHICH PROBLEM 
While edge offloading is very promising, its feasibility depends on the robustness of the wireless communication link~(\cite{L4S-ieee}). Robot applications typically include continuous, high bitrate uplink streams from sensors to the edge. 
When the bitrate exceeds available transmit capacity, queue buildup occurs, leading to increased latency and potentially packet loss.
These communication imperfections are not only network layer effects, but they also directly affect robot performance. Greater end-to-end delay diminishes control responsiveness, perception accuracy, and task reliability. For instance, delayed LiDAR updates can lead to localization drift~(\cite{damigos2023comaware}) or missed obstacle detections in-time~(\cite{martins2024impact}), having a direct impact on mission success.

%% HOW IS IT ADDRESSED 
To provide real-time response, the data to be transmitted must have a bitrate below the capacity of the wireless link for both uplink and downlink. In robotics, the focus is predominantly on the uplink, due to the volume of sensor data sent from the robot to the edge and the typical asymmetry between uplink and downlink capacity in cellular networks --- cellular networks are usually optimized to serve higher data rates in downlink.
Rate adaptation algorithms are required to adhere to these constraints. They must dynamically adapt the bitrate of outgoing data streams based on real-time network feedback in the attempt to prevent congestion and queuing. These algorithms, when properly used, ensure that the sensor data resolution is as high as possible while keeping network latency low.
However, most of the existing rate adaptation schemes have been developed for multimedia services~(\cite{steffens2021robustness}), notably video streaming, and are tailored towards perceptual quality metrics such as the Peak Signal-to-Noise Ratio (PSNR) and the Structural Similarity Index Measure (SSIM). Those do not match robotic KPIs such as distance errors, localization drift, or accurate object detection. Furthermore, 3D LiDAR data has a unique structure and data rate profile and requires different encoding approaches than traditional video.

In this paper, we extend the rate-adaptation \texttt{SCReAM v2} L4S-enabled framework~(\cite{johansson2024screamv2}) to support real-time uplink transmission of 3D LiDAR data for robotics. To do so, an appropriate point cloud codec is utilized, capable of adhering to the necessary real-time adjustment of the encoded bitrate, such that the channel capacity is respected. The key idea is connecting the real-time, lossy compression encoding process to an error signal, responsible for capturing the accuracy required for the successful execution of a specific task. Hence, enabling the framework to achieve low-latency, low-loss transmission of 3D LiDAR data to the edge under confined errors.
This makes it possible to offload computationally intensive LiDAR-based perception work to the edge and benefit from the real-time system performance. However, the proposed approach is not limited to perception. Other latency-sensitive applications, such as teleoperation, multi-agent coordination, or immersive VR/XR interaction, where high throughput sensor feedback is required, can also benefit.

%% WHAT IS ACHIEVED
This paper marks three main contributions. First, we establish a quantitative relationship between the compression error of point cloud encoders and the corresponding transmission bitrate, enabling real-time streaming of 3D LiDAR data under bounded reconstruction error. Second, we integrate real-time point cloud compression with a state-of-the-art congestion control and rate adaptation framework (L4S-enabled \texttt{SCReAM v2}), allowing low-latency high-bitrate uplink transmission over wireless networks to support edge offloading in robotics. Third, we validate the proposed system through field experiments and provide an open-source implementation to facilitate reproducibility and further research\footnote{Code and documentation available at: \magenta{\url{https://github.com/BBooda/SCReAM-LiDAR}}}.

%%%%%%%%%%%%%%%%%%%%%%%%%%%%%%%%% Related Work %%%%%%%%%%%%%%%%%%%%%%%%%%%%%%%%%
\section{Related Work} \label{sec:related-work}
%%%%%%%%%%%%%%%%%%%%%%%%%%%%%%%%%%%%%%%%%%%%%%%%%%%%%%%%%%%%%%%%%%%%%%%%%%%%%%%%

Robotic systems can benefit significantly from edge computing~\cite{seisa2022med}. The low latency and high computational capacity of edge computing can enable robotic applications such as single~\cite{dechouniotis2022edge, sarker2019offloading} and multi-robot SLAM~\cite{huang2022edge}, substantially reducing processing time. Similarly, offloading LiDAR data to edge servers can support computationally intensive real-time 3D segmentation~\cite{mclean2022towards}, obstacle~\cite{chen2022construction} and object~\cite{aung2024review} detection, as well as LiDAR-based perception~\cite{hou2025enhancing} and high-precision map generation~\cite{lee2020design} for autonomous driving. However, to enable such edge-assisted robotic applications, it is crucial to establish reliable real-time point cloud transmission.

Rate-adaptive algorithms for low-latency transmission over wireless networks have been a cornerstone of time-critical communications. A strong focus has been placed on algorithms such as \texttt{Google Congestion Control} (\cite{carlucci2016analysis}), which leverages queue delay gradient information to detect congestion and adapt the transmission rate. It has been primarily developed for video streaming. Similarly, targeting real-time video streaming, other state-of-the-art approaches include algorithms such as \texttt{SCReAM v2} (\cite{johansson2024screamv2}) and its L4S capabilities discussed herein. However, these algorithms do not address 3D LiDAR streaming and do not consider connections to robotics KPIs. At the same time, LiDAR-specific streaming frameworks have also been proposed. For instance, \cite{tu2019real} developed a 3D LiDAR streaming algorithm that focused mainly on the data encoder, without taking into account real-time communication requirements or channel variations. In contrast, \cite{anand2022novel} presented a novel 3D LiDAR framework that integrated state-of-the-art communication streaming technologies such as WebRTC and Firebase, although little attention was given to real-time adaptive compression of the data. More recently, \cite{Cao-real-time-icra} proposed a framework that primarily targets 3D point cloud encoder development while also adjusting the transmission bitrate according to the buffer queue length. This represents a notable step toward producing complete transmission pipelines that resemble those found in the video transmission literature. However, it is well known that real-time streaming over complex public networks requires additional mechanisms to ensure that network conditions are identified promptly and low-latency communication is achieved.

Finally, point cloud compression techniques range from standardized geometric encoders to recent learning-based approaches tailored for robotics. Widely used libraries such as \texttt{Draco} by~\cite{google_draco} and the \texttt{MPEG G-PCC} standard employ spatial partitioning to efficiently compress static geometry, offering reliable performance and broad compatibility.
For real-time robotic applications, learned compression schemes have demonstrated superior adaptability and efficiency.
% \cite{Theis2022riddle} compresses 3D LiDAR data by projecting it into range images and applying deep delta encoding to model local structural variations, significantly improving rate–distortion performance. 
Building on sparse representations,~\cite{you2025reno} introduced a real-time neural encoder based on sparse tensors, enabling high compression efficiency with low inference latency.
Finally and most recently,~\cite{stathoulopoulos2025sgadpcc} proposed a scene graph-aware compression pipeline that integrates semantics and scene structure into the encoding process, highlighting the trend towards task-aware compression for edge-assisted robotics.

%%%%%%%%%%%%%%%%%%%%%%%%%%%%%%%%%%%%%%%%%%%%%%%%%%%%%%%%%%%%%%%%%%%%%%%%%%%%%%%%
%%%%%%%%%%%%%%%%%%%%%%%%%%%%% Problem Formulation %%%%%%%%%%%%%%%%%%%%%%%%%%%%%%
\section{Preliminaries}\label{sec:preliminaries}
%%%%%%%%%%%%%%%%%%%%%%%%%%%%%%%%%%%%%%%%%%%%%%%%%%%%%%%%%%%%%%%%%%%%%%%%%%%%%%%%

To ground our contribution, we briefly outline the building blocks. We build on the L4S-enabled \texttt{SCReAM v2}, originally developed and standardized for rate-adaptive multimedia transmission over wireless networks by~\cite{johansson2018adaptive} and~\cite{L4S-ieee}. \texttt{SCReAM v2} has shown strong real-time performance in video, including vehicle teleoperation~(\cite{ericsson2023teledriving}) and edge offloading of vision-based perception (e.g., ORB-SLAM3 demonstrated by~\cite{olofsson2024l4s}). However, it does not natively support 3D LiDAR streaming, an underexplored area. We extend \texttt{SCReAM v2} with LiDAR-aware encoding and rate adaptation, ensuring transmissions respecting both network constraints and error bounds produced by data encoding. This framework lays the groundwork for low-latency, error-bounded delivery of 3D sensor data for robotics and other time-critical, edge-assisted applications.
%%%%%%%%%%%%%%%%%%%%%%%%%%%%%%%%%%%%%%%%%%%%%%%%%%%%%%%%%%%%%%%%%%%%%%%%%%%%%%%%
\begin{figure*}[ht]
    \hspace{-5mm}
    \includegraphics[width=\textwidth]{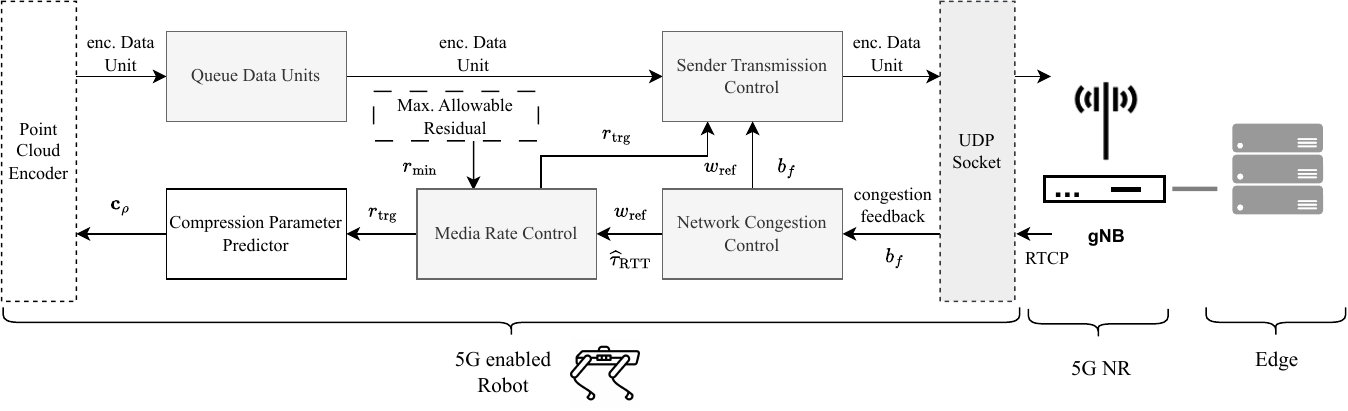}
    \caption{Sender transmission pipeline. Gray colored boxes depict the underline framework components, while the rest depict expansions native to the proposed methodology. The receiving side (not shown here) is responsible for rendering the 3D LiDAR data and generating the RTCP feedback.} 
    \label{fig:architecture}
\end{figure*}
%%%%%%%%%%%%%%%%%%%%%%%%%%%%%%%%%%%%%%%%%%%%%%%%%%%%%%%%%%%%%%%%%%%%%%%%%%%%%%%%

%%%%%%%%%%%%%%%%%%%%%%%%%%%%%%%%%%%%%%%%%%%%%%%%%%%%%%%%%%%%%%%%%%%%%%%%%%%%%%%%
\subsection{L4S-enabled \texttt{SCReAM v2} in a nutshell}
\label{network-congestion-control}
%%%%%%%%%%%%%%%%%%%%%%%%%%%%%%%%%%%%%%%%%%%%%%%%%%%%%%%%%%%%%%%%%%%%%%%%%%%%%%%%
At a high level, L4S-enabled \texttt{SCReAM v2} regulates the sender via a transmission controller that forwards encoded data units over UDP while bounding the bytes in flight (unacknowledged bytes) against a reference window. The bound is soft: a configurable overshoot margin permits brief exceedance of the reference to avoid unnecessary queuing delay. A pacing rate, driven by the media-rate controller’s target bitrate, smooths transmissions so large bursts do not hit the wireless link, reducing the risk of queue overflows and burst-related issues. As illustrated in Fig.~\ref{fig:architecture}, gray boxes denote native \texttt{SCReAM v2} components and the rest indicate our extensions which will be discussed later on. The sender window is formulated as:
%%%%%%%%%%%%%%%%%%%%%%%%%%%%%%%%%%%%%%%%%%%%%%%%%%%%%%%%%%%%%%%%%%%%%%%%%%%%%%%%
\begin{equation}
    w_s = w_\text{ref} \cdot w_\text{ref}^\text{over} \cdot F_s - b_{f},
\end{equation}
%%%%%%%%%%%%%%%%%%%%%%%%%%%%%%%%%%%%%%%%%%%%%%%%%%%%%%%%%%%%%%%%%%%%%%%%%%%%%%%%
where, $w_s$ is the sender window (bytes); $w_\text{ref}$ is the reference window; $w_\text{ref}^\text{over}$ is the permitted overshoot factor; $F_s$ is the data-unit frame size (bytes); and  $b_f$ is the current bytes in flight.
The Real-time Transport Control Protocol (RTCP) feedback supplies the congestion controller with received-byte counters and timing metadata to estimate congestion and queuing delay. The controller computes the reference window and passes it, along with the current bytes in flight, to the sender’s transmission controller. When congestion is detected, the reference window is multiplicatively reduced as:
%%%%%%%%%%%%%%%%%%%%%%%%%%%%%%%%%%%%%%%%%%%%%%%%%%%%%%%%%%%%%%%%%%%%%%%%%%%%%%%%
\begin{equation}
    w_\text{ref} = \alpha_\text{loss} \cdot (\beta_\text{ecn} \lor \beta_\text{l4s}) \cdot q_\text{trg} \cdot w_\text{ref}.
\end{equation}
%%%%%%%%%%%%%%%%%%%%%%%%%%%%%%%%%%%%%%%%%%%%%%%%%%%%%%%%%%%%%%%%%%%%%%%%%%%%%%%%
Here, $\alpha_\text{loss}$ captures loss-based reductions; $\beta_\text{ecn}$ or $\beta_\text{l4s}$ apply ECN-based scaling depending on Classic ECN or L4S mode; and $q_\text{trg}$ weights the response relative to the queue delay target.
The reference window is also provided to the media-rate controller, which maps it to a target bitrate and updates whenever $w_\text{ref}$ changes. Stabilizing safeguards are applied when the window is very small or when L4S feedback is sparse, improving robustness. Frequent congestion notifications enable faster reaction to capacity shifts, allowing queues to drain promptly. The target bitrate is computed as:
%%%%%%%%%%%%%%%%%%%%%%%%%%%%%%%%%%%%%%%%%%%%%%%%%%%%%%%%%%%%%%%%%%%%%%%%%%%%%%%%
\begin{equation} \label{eq:target_bitrate}
    r_\text{trg} = \frac{8 \cdot w_\text{ref}}{\widehat{\tau}_\text{RTT}} 
\end{equation}
%%%%%%%%%%%%%%%%%%%%%%%%%%%%%%%%%%%%%%%%%%%%%%%%%%%%%%%%%%%%%%%%%%%%%%%%%%%%%%%%
where, $r_\text{trg}$ is the target bitrate and $\widehat{\tau}_\text{RTT}$ is the smoothed round-trip time (RTT), obtained by low-pass filtering RTP transmit/receive delays.

%%%%%%%%%%%%%%%%%%%%%%%%%%%%%%%%%%%%%%%%%%%%%%%%%%%%%%%%%%%%%%%%%%%%%%%%%%%%%%%%
\subsection{Problem Formulation} \label{problem_formulation}
%%%%%%%%%%%%%%%%%%%%%%%%%%%%%%%%%%%%%%%%%%%%%%%%%%%%%%%%%%%%%%%%%%%%%%%%%%%%%%%%

In practice, our method targets two objectives: (i) real-time delivery of 3D LiDAR data from a 5G-equipped robot to an edge server, and (ii) error-bounded compression whose artifacts remain small enough to preserve application-level performance.
%%%%%%%%%%%%%%%%%%%%%%%%%%%%%%%%%%%%%%%%%%%%%%%%%%%%%%%%%%%%%%%%%%%%%%%%%%%%%%%%
\subsubsection{1) Data encoding at the required bitrate:}
%%%%%%%%%%%%%%%%%%%%%%%%%%%%%%%%%%%%%%%%%%%%%%%%%%%%%%%%%%%%%%%%%%%%%%%%%%%%%%%%

The first objective is met by equipping \texttt{SCReAM v2} with a LiDAR-compatible encoder module that can track the network-driven target bitrate, keeping the application-side sender queue draining (or at least non-growing) to enable real-time transmission. We assume the encoder can produce encoded data units at the requested rate $r_\text{trg}$ (bps). Traditional media stacks already support this, for example, \texttt{H.264} over \texttt{GStreamer}~(\cite{gstreamer}). For real-time LiDAR data, prior work offers multiple compression algorithms, most of which expose tunable parameters (e.g., quantization bits, compression level, etc.) that can be set to compress a scan or a series of scans over time. To remain encoder-agnostic and support our goal of a generic real-time LiDAR compression framework, we define a mapping between the target bitrate and the encoder configuration as follows.

Let the compression parameter vector be defined as $\mathbf{c}_\rho = (d_1, d_2, \ldots, d_{D_\rho}) \in \mathcal{C}_\rho$, where $\mathcal{C}_\rho \subseteq \mathbb{R}^{D_\rho}$ is the configuration space of compression algorithm $\rho$ and $D_\rho$ is its number of tunable parameters. Each component of $\mathbf{c}_\rho$ controls an encoding knob (e.g., quantization level, spatial resolution, predictive mode). Let $P_k = \{(x_i,y_i,z_i)^\top \in \mathbb{R}^3\}_{i=1}^{N}$ denote the $k$-th scan, encoded using $\mathbf{c}_\rho$. We assume $|P_k|=N$ is fixed for a given LiDAR sensor, with padding applied if needed. Then, we define $f$ as the mapping from configuration, given the scan cardinality, to the achieved (average) bitrate in bits per second when encoding a sequence of scans:
%%%%%%%%%%%%%%%%%%%%%%%%%%%%%%%%%%%%%%%%%%%%%%%%%%%%%%%%%%%%%%%%%%%%%%%%%%%%%%%%
\begin{equation}\label{eq:generic-predicted-bps}
f(\mathbf{c}_\rho\,|\,N) = \hat{r},\quad f:\ \mathcal{C}_\rho \times \mathbb{N} \to \mathbb{R}^{+},
\end{equation}
%%%%%%%%%%%%%%%%%%%%%%%%%%%%%%%%%%%%%%%%%%%%%%%%%%%%%%%%%%%%%%%%%%%%%%%%%%%%%%%%
where $r$ is the predicted bitrate (bps).

For a fixed scanning frequency, the mapping $f(\cdot)$ can be obtained analytically or via a learning-based model. In principle, one could invert Eq.~\eqref{eq:generic-predicted-bps} and, given the target bitrate $r_{\text{trg}}$ from \texttt{SCReAM v2}, solve for $\mathbf{c}_\rho$. However, $f$ is not strictly invertible with respect to $\mathbf{c}_\rho$ because the achieved bitrate depends jointly on $\mathbf{c}_\rho$ and the scan geometry (e.g., quantization and spatial clustering effects). 
In practice, we therefore interpret $f^{-1}(\hat{r} \mid N)$ as an optimization step that selects $\mathbf{c}_\rho$ whose induced bitrate approximates the target.

%%%%%%%%%%%%%%%%%%%%%%%%%%%%%%%%%%%%%%%%%%%%%%%%%%%%%%%%%%%%%%%%%%%%%%%%%%%%%%%%
\subsubsection{2) Confining Compression Residual:}
%%%%%%%%%%%%%%%%%%%%%%%%%%%%%%%%%%%%%%%%%%%%%%%%%%%%%%%%%%%%%%%%%%%%%%%%%%%%%%%%
The second objective is achieved by mapping compression artifacts to a distortion metric (residual) perceivable by robotic applications. In this work, we consider lossy compression; once a scan is compressed and decompressed, the result will not perfectly match the original. We can capture this difference in various ways; for simplicity, we utilize the Euclidean ($\ell_2$) norm $\|\Delta P_k\|_2$. However, other operators that best fit the chosen encoder can also be used, such as the \textit{Chamfer Distance} if the encoder does not keep the original cardinality $N$ of the scan. Overall, we aim to keep this residual small. Let:
%%%%%%%%%%%%%%%%%%%%%%%%%%%%%%%%%%%%%%%%%%%%%%%%%%%%%%%%%%%%%%%%%%%%%%%%%%%%%%%%
\begin{equation}
g_\rho(P_k, \mathbf{c}_\rho) = \Delta P_k, \label{eq:residual-generic}
\end{equation}
%%%%%%%%%%%%%%%%%%%%%%%%%%%%%%%%%%%%%%%%%%%%%%%%%%%%%%%%%%%%%%%%%%%%%%%%%%%%%%%%
where $g_\rho(\cdot)$ denotes the residual function of the encoder $\rho$. In our case, the residual arises from the \texttt{Draco} encoder~(\cite{google_draco}), and the compression parameter vector $\mathbf{c}_\rho$.
Given that the decompressed scan and the original have the same number of points (handled by \texttt{Draco}), we proceed to define the following residual. Let the compressed point cloud be denoted by:
%%%%%%%%%%%%%%%%%%%%%%%%%%%%%%%%%%%%%%%%%%%%%%%%%%%%%%%%%%%%%%%%%%%%%%%%%%%%%%%%
\begin{equation}
\hat{P}_k = \{(x_i, y_i, z_i)^\top + (\delta^x_i, \delta^y_i, \delta^z_i)^\top\}_{i=1}^N \in \mathbb{R}^3
\end{equation}
%%%%%%%%%%%%%%%%%%%%%%%%%%%%%%%%%%%%%%%%%%%%%%%%%%%%%%%%%%%%%%%%%%%%%%%%%%%%%%%%
Their difference, representing the compression artifact or distortion, can then be defined as $\hat{P}_k - P_k = \Delta P_k$.
For a given application and at a given time, we wish to select $\mathbf{c}_\rho$ such that:
%%%%%%%%%%%%%%%%%%%%%%%%%%%%%%%%%%%%%%%%%%%%%%%%%%%%%%%%%%%%%%%%%%%%%%%%%%%%%%%%
\begin{equation} \label{eq:compression-error-limit}
\big\|g_\rho(P_k, \mathbf{c}_\rho)\big\| \leq \varepsilon
\end{equation}
%%%%%%%%%%%%%%%%%%%%%%%%%%%%%%%%%%%%%%%%%%%%%%%%%%%%%%%%%%%%%%%%%%%%%%%%%%%%%%%%
where $\varepsilon$ represents a maximum acceptable critical distortion threshold, and $g(\cdot)$ represents the chosen function representing the compression artifacts, in our case $\Delta P_k$.
Connecting to the target bitrate from the transmission framework, we can select a corresponding $\mathbf{c}_\rho$ that respects Eq.~\eqref{eq:compression-error-limit} and can be linked to a corresponding bitrate. This particular bitrate can define the lower critical limit $r_{\text{min}}$ that our rate-adaptive framework can operate with. Such an analysis can be applied as an offline optimization configuration step, aiming to make the particular application use-case robust.
%%%%%%%%%%%%%%%%%%%%%%%%%%%%%%%%%%%%%%%%%%%%%%%%%%%%%%%%%%%%%%%%%%%%%%%%%%%%%%%%
\begin{figure}[!b]
    \centering
    \begin{subfigure}[t]{0.49\linewidth} 
        \centering
        \includegraphics[width=\linewidth]{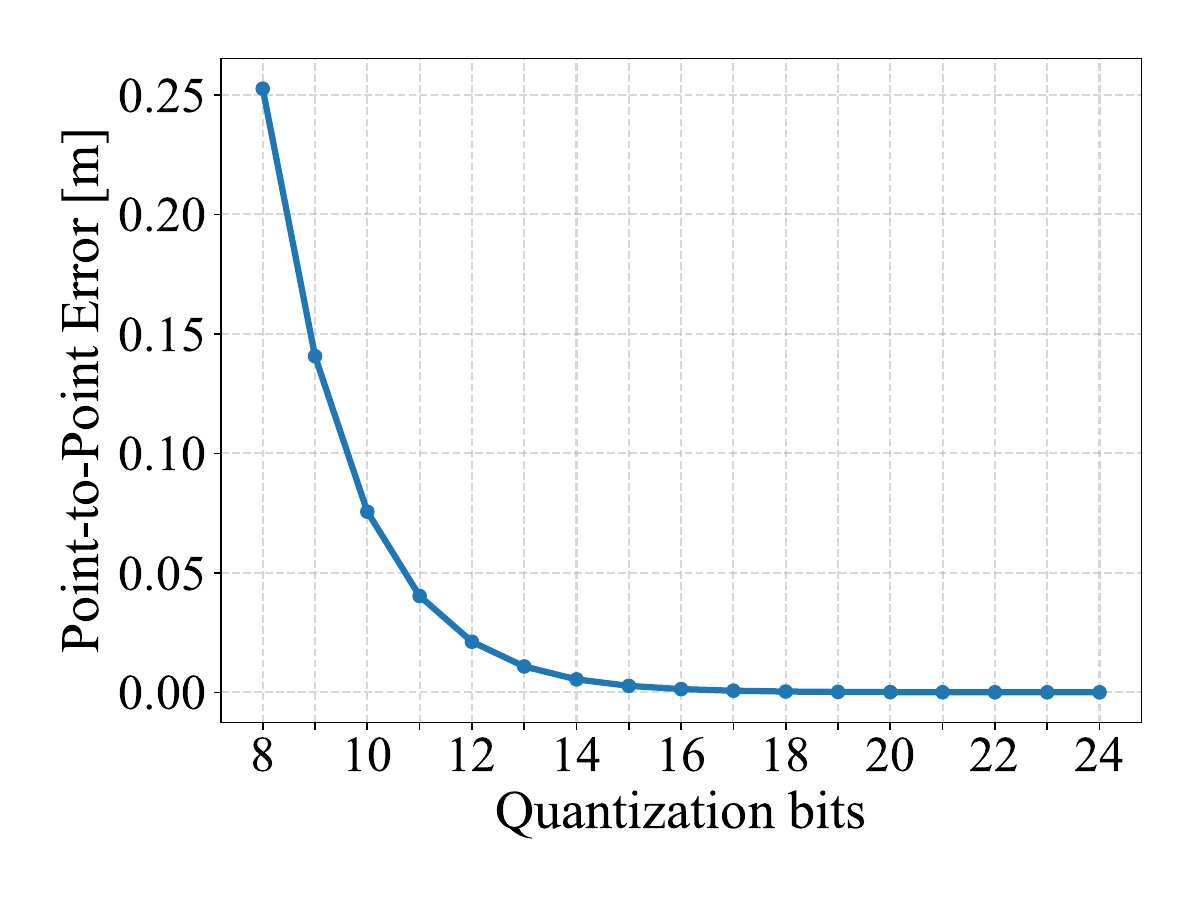}
        % \caption{Mean PtP error vs. quantization bits.}
        % \label{fig:prel:ptp}
    \end{subfigure}
    \hfill
    \begin{subfigure}[t]{0.49\linewidth}
        \centering
        \includegraphics[width=\linewidth]{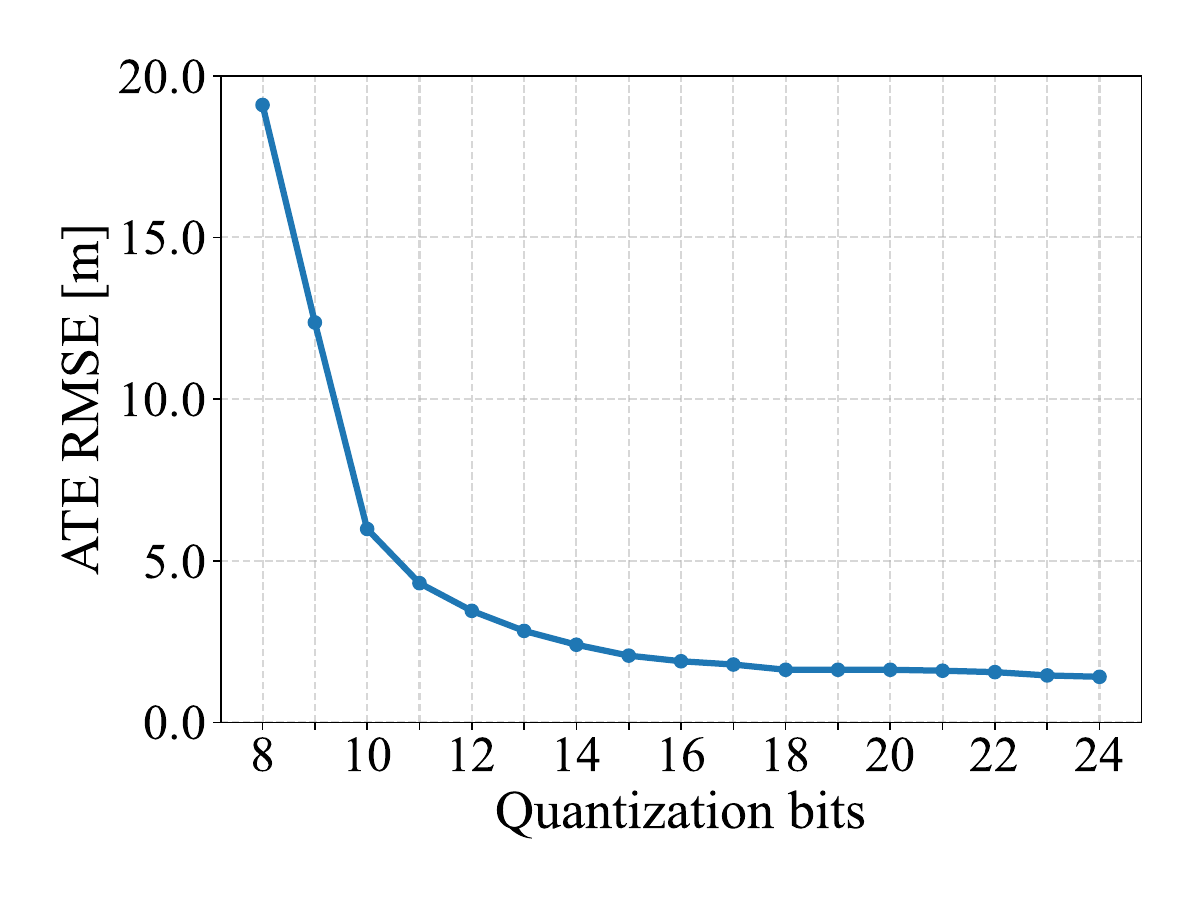}
        % \caption{Trajectory ATE RMSE error vs. quantization bits.}
        % \label{fig:prel:ate}
    \end{subfigure}
    \vspace{-0.8cm}
    \caption{Left: mean point-to-point error between raw LiDAR scans and their \texttt{Draco}-compressed counterparts as a function of quantization bits $q$. Right: Absolute Trajectory Error (ATE, RMSE) of \texttt{KISS-SLAM} versus $q$ for the same sweep. The ground-truth trajectory is obtained by running \texttt{KISS-SLAM} on the uncompressed data, while the compressed runs are evaluated against this reference.} \label{fig:prel:errors}
\end{figure}
%%%%%%%%%%%%%%%%%%%%%%%%%%%%%%%%%%%%%%%%%%%%%%%%%%%%%%%%%%%%%%%%%%%%%%%%%%%%%%%%

%%%%%%%%%%%%%%%%%%%%%%%%%%%%%%%%%%%%%%%%%%%%%%%%%%%%%%%%%%%%%%%%%%%%%%%%%%%%%%%%
\subsubsection{3) Minimum $\varepsilon$ per considered application:}
%%%%%%%%%%%%%%%%%%%%%%%%%%%%%%%%%%%%%%%%%%%%%%%%%%%%%%%%%%%%%%%%%%%%%%%%%%%%%%%%
Critical applications often degrade under aggressive compression~(\cite{martins2024impact,wang2023robust,zhang2024impact,steffens2021robustness,tu2019real,stathoulopoulos2025sgadpcc}). For stateful robotic workloads, such as object detection, vision–language models (VLMs), and SLAM, bounding compression error helps maintain required performance.

To illustrate this effect, Fig.~\ref{fig:prel:errors} plots the mean point-to-point error versus the \texttt{Draco} quantization bits $q$. Here, $q$ is the dominant parameter in \texttt{Draco}'s compression vector, $\mathbf{c}_{\texttt{Draco}}$. Then, Fig.~\ref{fig:prel:errors} also reports the Absolute Trajectory Error (ATE) RMSE of \texttt{KISS-SLAM}~(\cite{guadagnino2025kiss}) as a function of \(q\). For these experiments, we disable the transmission framework (fully onboard execution) to isolate and evaluate the compression effects. The ground truth is obtained by running the same SLAM algorithm on the uncompressed LiDAR data.
% For example, $\|g(p, \mathbf{c}^{r})\| \leq \varepsilon^c$, where $\varepsilon^c$ defines the application lower limit. 
% In practice, $\varepsilon^c$ can vary over time and depends on the specific application itself. This can be attributed to the fact that some applications experience different challenges at different times, for example, it has been documented that SLAM may diverge from the ground truth when traversing from indoors to outdoors. Another scenario can relate to different mission requirements at different times. For example, throughout a mission, a robotic platform can traverse open areas where collision risk is less likely, and cluttered space where critical infrastructure or humans are present. The second case is obviously more critical. Nevertheless, since the focus on the research does not lie on the specific applications, rather in the communication framework, we are going to consider one static $\varepsilon^c$. Further $\varepsilon^c$ can be defined either analytically or experimentally. 

%%%%%%%%%%%%%%%%%%%%%%%%%%%%%%%%%%%%%%%%%%%%%%%%%%%%%%%%%%%%%%%%%%%%%%%%%%%%%%%%
%%%%%%%%%%%%%%%%%%%%%%%%%%%%% System Architecture %%%%%%%%%%%%%%%%%%%%%%%%%%%%%%
\section{The Proposed System Architecture}\label{System_architecture}
%%%%%%%%%%%%%%%%%%%%%%%%%%%%%%%%%%%%%%%%%%%%%%%%%%%%%%%%%%%%%%%%%%%%%%%%%%%%%%%%
To complete the transmission pipeline, our added components must meet the performance requirements stated in the problem formulation. We introduce the following modules to adapt L4S-enabled \texttt{SCReAM v2} to accommodate 3D LiDAR data: the \textit{Compression Parameter Predictor}, which selects encoder controls to track the target bitrate; we incorporate the \textit{Point Cloud Encoder}, which produces encoded data units at the commanded rate so the \textit{Queued Data Units} drain quickly; and the \textit{Residual Error Optimizer} module, which links the application’s error budget $\varepsilon$ to a minimum allowable transmission target bitrate via an offline optimization (learned offline, deployed online).
If the encoder’s output rate exceeds the target, data units accumulate and application-layer latency grows; if it falls well below the target, compression is unnecessarily aggressive and the risk of violating the critical error $\varepsilon$ increases. Therefore, encoder parameters should be chosen so that the produced bitrate remains as close as possible to the target. The proposed modules and their interconnections are shown in Fig.~\ref{fig:architecture}.

%%%%%%%%%%%%%%%%%%%%%%%% Low-Latency Data Transmission %%%%%%%%%%%%%%%%%%%%%%%%%
\subsection{Compression Parameter Predictor} \label{sec:modeling-draco-bps}
%%%%%%%%%%%%%%%%%%%%%%%%%%%%%%%%%%%%%%%%%%%%%%%%%%%%%%%%%%%%%%%%%%%%%%%%%%%%%%%%
In our implementation we use \texttt{Draco}, which lacks an official analytical model. Our goal is to make the encoder produce a bitrate that tracks the congestion-controlled target in Eq.~\eqref{eq:target_bitrate}. To model Eq.~\eqref{eq:generic-predicted-bps} on the encoder side, we adopt a learning-based predictor that generalizes to other encoders while respecting channel capacity. For \texttt{Draco}, the encoding controls are:
%%%%%%%%%%%%%%%%%%%%%%%%%%%%%%%%%%%%%%%%%%%%%%%%%%%%%%%%%%%%%%%%%%%%%%%%%%%%%%%%
\begin{align}
    & q \in Q = \{q \in \mathbb{Z}^+: 8 \le q \le 24 \} 
    \quad \text{(quantization bits)}, \label{eq:q-param-draco} \\
    & c \in C = \{c \in \mathbb{Z}^+_0: 0 \le c \le 9 \}
    \quad \text{(compression level)}. \label{eq:c-param-draco}
\end{align}
%%%%%%%%%%%%%%%%%%%%%%%%%%%%%%%%%%%%%%%%%%%%%%%%%%%%%%%%%%%%%%%%%%%%%%%%%%%%%%%%
We use a second-order polynomial map $\Phi:\mathbb{R}^3 \to \mathbb{R}^9$ and a linear model with parameters $\alpha \in \mathbb{R}^9$ and $\beta \in \mathbb{R}$. For any $(q,c)$ and number of points $N$, the bitrate is modeled as:
%%%%%%%%%%%%%%%%%%%%%%%%%%%%%%%%%%%%%%%%%%%%%%%%%%%%%%%%%%%%%%%%%%%%%%%%%%%%%%%%
\begin{equation}
    \hat{r} \,=\, f(q, c \, | \, N) \,\approx\, \alpha^\top \Phi(q,c,N) \,+\, \beta,
\end{equation}
%%%%%%%%%%%%%%%%%%%%%%%%%%%%%%%%%%%%%%%%%%%%%%%%%%%%%%%%%%%%%%%%%%%%%%%%%%%%%%%%
where $\hat{r}$ is the predicted bitrate (bps) from $f$ in Eq.~\eqref{eq:generic-predicted-bps}, given $(q,c,N)$.

To infer the compression parameters, we invert the mapping as follows. Define the parameter grid:
%%%%%%%%%%%%%%%%%%%%%%%%%%%%%%%%%%%%%%%%%%%%%%%%%%%%%%%%%%%%%%%%%%%%%%%%%%%%%%%%
\begin{equation}
    \mathcal{G} \;=\; Q \times C \;=\; \{(q_i,c_j)\},
\end{equation}
%%%%%%%%%%%%%%%%%%%%%%%%%%%%%%%%%%%%%%%%%%%%%%%%%%%%%%%%%%%%%%%%%%%%%%%%%%%%%%%%
and, for a fixed number of points \(N = |P_k|\) (same sensor), evaluate the predicted rate at each grid point:
%%%%%%%%%%%%%%%%%%%%%%%%%%%%%%%%%%%%%%%%%%%%%%%%%%%%%%%%%%%%%%%%%%%%%%%%%%%%%%%%
\begin{equation}
    \hat{r}_{ij} \;\approx\; f(q_i,c_j\,|\,N), 
    \quad \forall (i,j)\ \text{with}\ (q_i,c_j) \in \mathcal{G}.
    \label{eq:pij-grid}
\end{equation}
%%%%%%%%%%%%%%%%%%%%%%%%%%%%%%%%%%%%%%%%%%%%%%%%%%%%%%%%%%%%%%%%%%%%%%%%%%%%%%%%
Given a target bitrate $r_\text{trg}$, we select the grid point whose predicted rate is closest to $r_\text{trg}$:
%%%%%%%%%%%%%%%%%%%%%%%%%%%%%%%%%%%%%%%%%%%%%%%%%%%%%%%%%%%%%%%%%%%%%%%%%%%%%%%%
\begin{align}
    (q^*, c^*) 
    &= \arg \min_{(q_i, c_j) \in \mathcal{G}} |\hat{r}_{ij} - r_\text{trg}|.
    \label{eq:predict-draco-params}
\end{align}
%%%%%%%%%%%%%%%%%%%%%%%%%%%%%%%%%%%%%%%%%%%%%%%%%%%%%%%%%%%%%%%%%%%%%%%%%%%%%%%%
We then apply these parameters in the \texttt{Draco} encoder, i.e., $\mathbf{c}_\rho^* = (q^*, c^*)$, to produce a bitrate that tracks the requested target, preventing queue growth and maintaining low latency. This selection is repeated for each new $r_\text{trg}$ reported by the transmission framework.

%%%%%%%%%%%%%%%%%%%%%%%%%%%%%%%%%%%%%%%%%%%%%%%%%%%%%%%%%%%%%%%%%%%%%%%%%%%%%%%%
\subsection{Residual Error Optimizer}
%%%%%%%%%%%%%%%%%%%%%%%%%%%%%%%%%%%%%%%%%%%%%%%%%%%%%%%%%%%%%%%%%%%%%%%%%%%%%%%%
Using the residual definition in Eq.~\eqref{eq:residual-generic}, the norm constraint in Eq.~\eqref{eq:compression-error-limit}, and the encoder grid $\mathcal{G}$ (with rates tabulated in Eq.~\eqref{eq:pij-grid}), we compute the minimum allowable bitrate $r_{\text{min}}$ that respects the critical residual bound $\varepsilon^c$.
%%%%%%%%%%%%%%%%%%%%%%%%%%%%%%%%%%%%%%%%%%%%%%%%%%%%%%%%%%%%%%%%%%%%%%%%%%%%%%%%
\begin{equation}
\label{eq:min_rate}
\begin{aligned}
    r_{\text{min}}
    &= \min_{(q_i, c_j) \in \mathcal{G}} \; f(q_i, c_j\,|\, N) \\
    &\text{s.t.}\;\; \big\|g_\rho(P_k, \mathbf{c}_\rho)\big\| \le \varepsilon,
    \;\; \text{with } \mathbf{c}_\rho = (q_i, c_j).
\end{aligned}
\end{equation}
%%%%%%%%%%%%%%%%%%%%%%%%%%%%%%%%%%%%%%%%%%%%%%%%%%%%%%%%%%%%%%%%%%%%%%%%%%%%%%%%
Practically, the trained model $f(q,c\,|\,N)$ is used both to predict the desired compression parameters $\mathbf{c}_\rho$ (via Eq.~\eqref{eq:predict-draco-params}) and, together with the residual constraint, to determine $r_{\text{min}}$. More generally, one could learn a separate surrogate for $g_\rho(P_k,\mathbf{c}_\rho)$ and use it to enforce the constraint directly, or incorporate the residual constraint into Eq.~\eqref{eq:predict-draco-params}. 

%%%%%%%%%%%%%%%%%%%%%%%%%%%%%%%%%%%%%%%%%%%%%%%%%%%%%%%%%%%%%%%%%%%%%%%%%%%%%%%%
%%%%%%%%%%%%%%%%%%%%%%%%%%% Experimental Validation %%%%%%%%%%%%%%%%%%%%%%%%%%%%
\section{Experimental Validation}
\label{sec:experimental_validation}
%%%%%%%%%%%%%%%%%%%%%%%%%%%%%%%%%%%%%%%%%%%%%%%%%%%%%%%%%%%%%%%%%%%%%%%%%%%%%%%%
The proposed framework has been extensively evaluated with hardware experiments in a public 5G network using a sensor kit including an \texttt{Ouster OS0-32} 3D LiDAR, and a \texttt{Quectel RM520N-GL} 5G modem. The sensor kit also includes an onboard processing unit \texttt{Intel Nuc}, and can be easily mounted in any vehicle or robotic platform with the corresponding form factor. 

The experiments took place in the area of Luleå University of Technology, in Luleå Sweden, and also included large-scale experiments involving multi-kilometer drives, with the speed of approximately $40-45 \, km/h$. The large-scale conditions, varying radio conditions, varying network conditions, and varying physical environment captured by the LiDAR sensor make this evaluation the first real-world rate-adaptive 3D LiDAR transmission framework evaluated in non-controlled network conditions. The proposed solution was able to achieve low-latency, and resilience in cell-edge and varying signal conditions, all while keeping the required minimum bitrare resulting in avoiding the violation of the maximum allowable error. The corresponding results are shown below, while the readers are also encouraged to watch the accompanying video.  
%%%%%%%%%%%%%%%%%%%%%%%%%%%%%%%%%%%%%%%%%%%%%%%%%%%%%%%%%%%%%%%%%%%%%%%%%%%%%%%%
\subsection{Framework evaluation}
%%%%%%%%%%%%%%%%%%%%%%%%%%%%%%%%%%%%%%%%%%%%%%%%%%%%%%%%%%%%%%%%%%%%%%%%%%%%%%%%
Regarding the performance of the proposed framework in terms of communication KPIs, we first investigate the congestion control mechanisms that result in the real-time transmission of data and the smooth integration of our encoding modules. 

\begin{figure}[ht]
    \includegraphics[width=\linewidth]{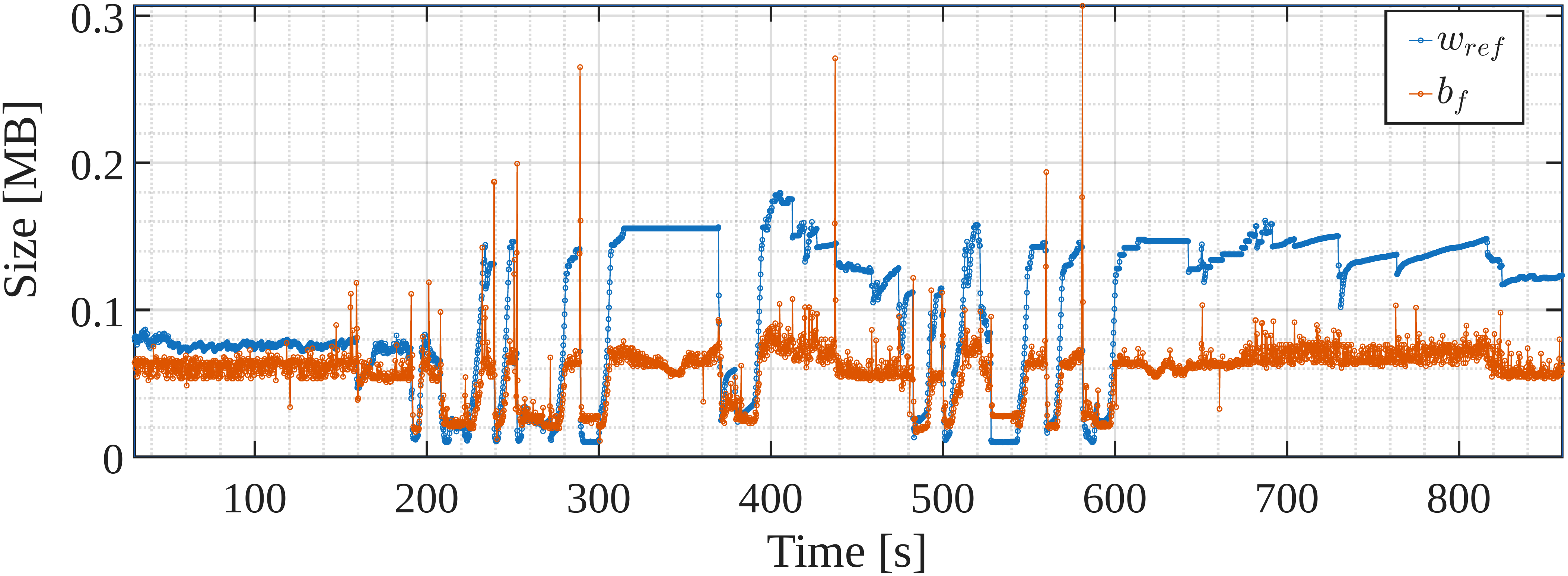}
    \caption{Reference window $w_{ref}$ vs. Bytes in Flight $b_f$.} 
    \label{fig:res:cwnd-vs-bif}
\end{figure}

Looking at the Sender Transmission Control, Figure \ref{fig:res:cwnd-vs-bif} shows the relationship between the Bytes in Flight (BIF) and the reference window $w_\text{ref}$ over time. Overall, $w_\text{ref}$ is timely updated and the BIF  are correctly controlled to closely follow $w_\text{ref}$ throughout all the varying conditions in this characteristic experiment. 

\begin{figure}[ht]
    \includegraphics[width=\linewidth]{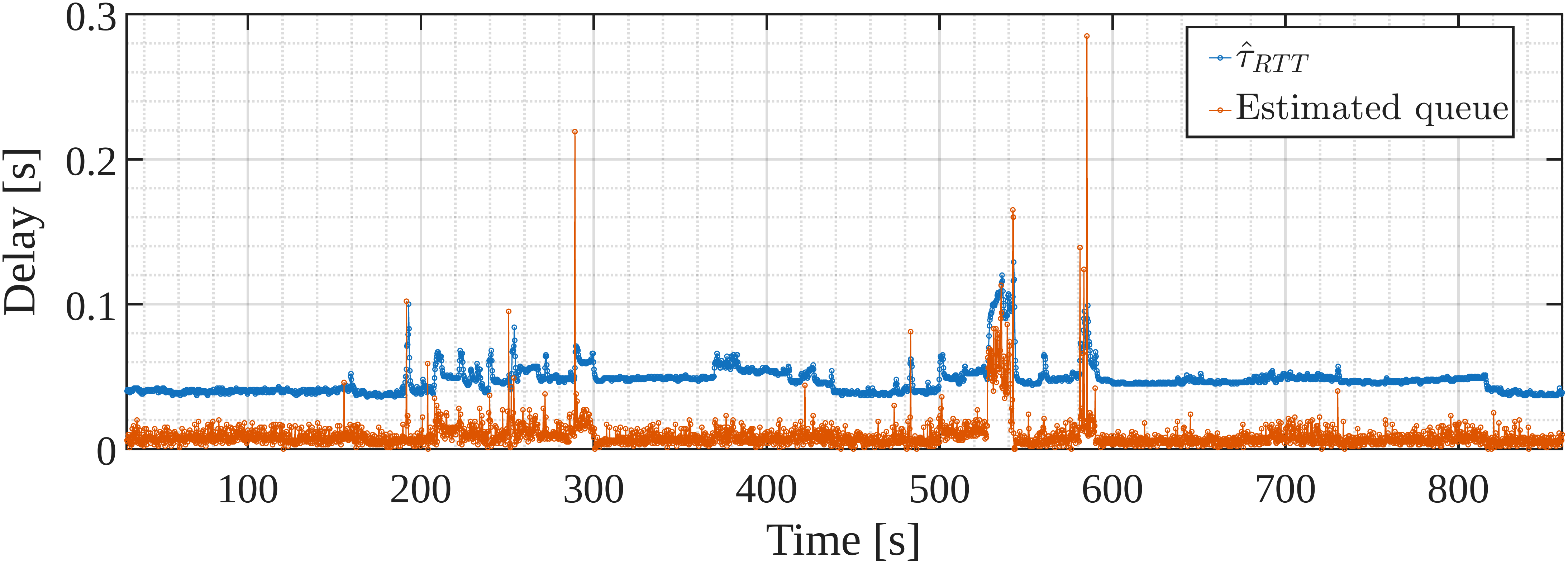}
    \caption{Smooth round trip time $\hat{\tau}_{RTT}$ vs. estimated queue delay.} 
    \label{fig:res:rtt-vs-eqd}
\end{figure}

\begin{figure}[ht]
    \centering
    \begin{subfigure}{\linewidth}
        \centering
        \includegraphics[width=1.0\linewidth]{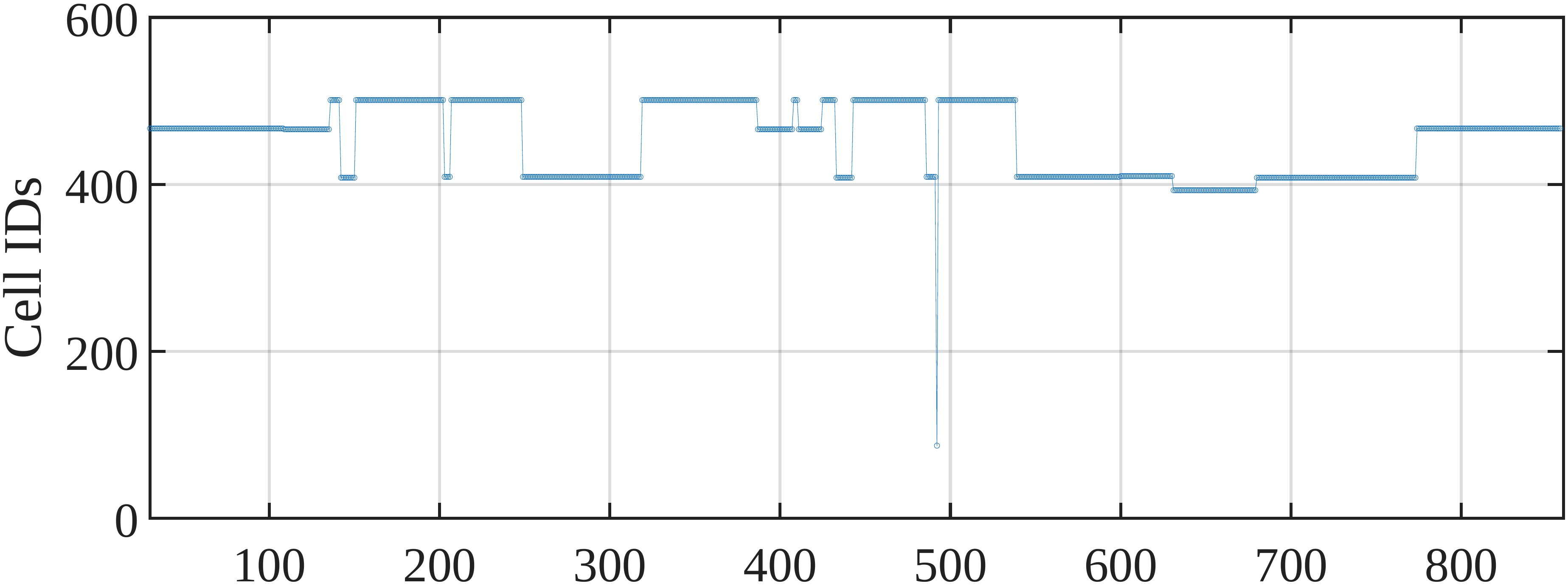}
        \caption{Handovers over time.}
        \label{fig:res:handovers}
    \end{subfigure}
    
    \vspace{0.5em} % small vertical space between plots

    \begin{subfigure}{1.0\linewidth}
        \centering
        \includegraphics[width=\linewidth]{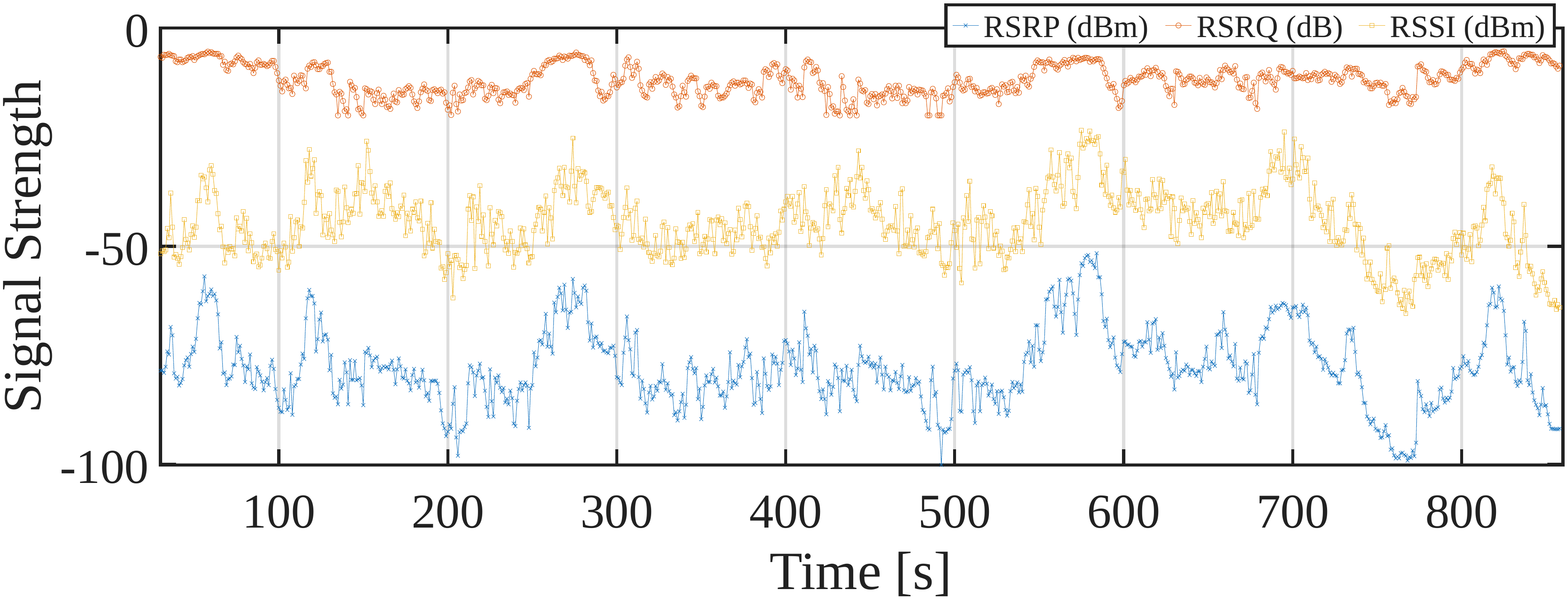}
        \caption{Radio signal KPIs over time.}
        \label{fig:res:radio-kpis}
    \end{subfigure}

    \caption{(a) shows how serving cells handovers are changing throughout the mission. (b) shows the radio signal KPIs throughout the mission, significantly low signal power is  experienced around $200$, $490$, and $780$ s. }
    \label{fig:res:lte-kpis}
\end{figure}

Of particular interest are the regions \textit{$180 - 300\, s$}, and \textit{$500 - 600\, s$}. Here the channel capacity notably drops capturing the absolute lowest limit of the $w_\text{ref}$. At the same time BIF are allowed to overcome the reference window. This is allowable until a certain point (slack) in \texttt{SCReAM v2} controlled by $w^\text{over}_\text{ref} = 5$, since it is beneficial in cases where large data chunks (large frames when dealing with video, or large LiDAR scans here) are transmitted and not unnecessarily queued up on the sender side, while still having the feedback that prevents a large network queue. This behavior, given the varying and at times quite challenging conditions (also shown in \ref{fig:res:radio-kpis}), result in the slight but bounded rise of RTT latency visible at the time-synchronized Figure \ref{fig:res:rtt-vs-eqd}. 

A further increase in latency is prevented by the fact that the amount that BIF can exceed the $w_\text{ref}$ size is capped at the shown ratio, while for all the remaining experiment, $w_\text{ref}$ values are always larger than BIF. At the same time, the encoded bitrate follows the target bitrate $r^t$ signal, and actually captures a slightly lower value throughout the depicted experiment. This contributes to the fact that RTT latency does not rise further since no extra data is queued when the channel capacity is low. 

\begin{figure}[ht]
    \includegraphics[width=\linewidth]{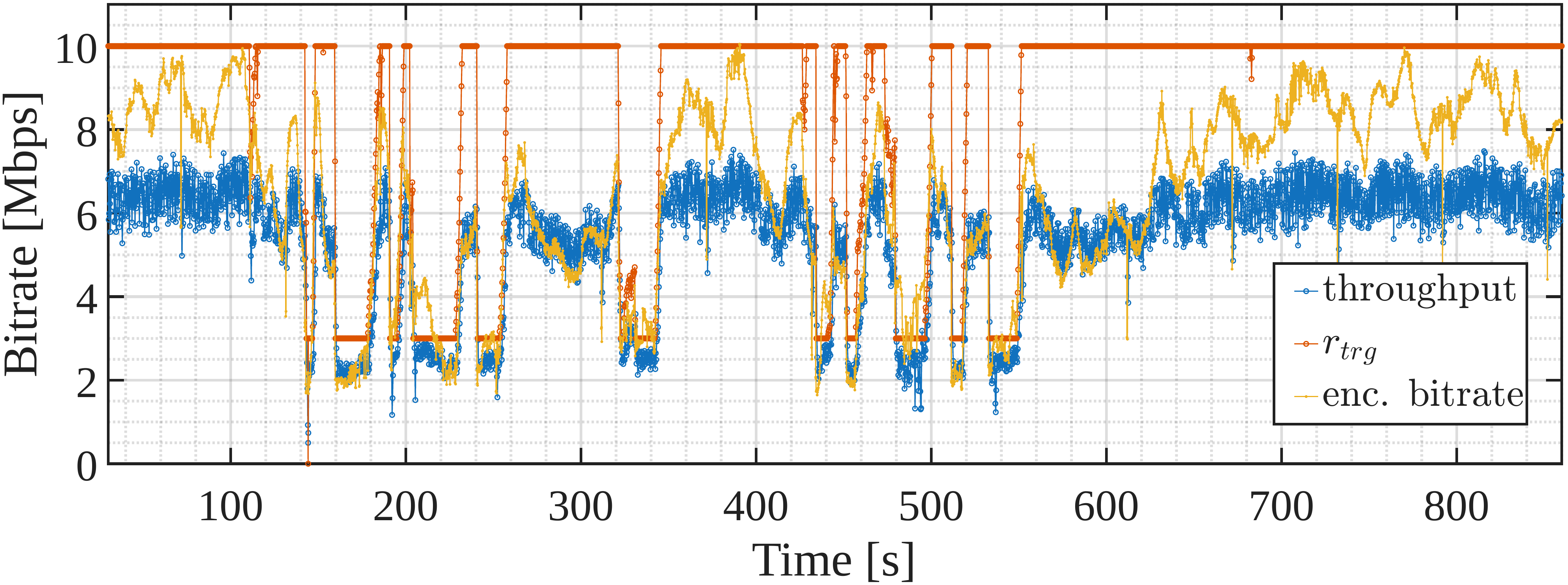}
    \caption{Framework throughput vs. target bitrate $r_\text{trg}$, vs. predicted encoder bitrate.} 
    \label{fig:res:th-vs-bitrate}
\end{figure}

The encoded bitrate is shown as enc. bitrate in Figure \ref{fig:res:th-vs-bitrate} which is the measured bitrate right after the Point Cloud Encoder module. Here we can also observe the performance of our proposed Compression Parameter Predictor (CPP). The performance of this module can be seen comparing the target bitrate and the enc. bitrate. The enc. bitrate closely follows the target bitrate $r_\text{trg}$ reference. It has to be noted that the target bitrate signal is allowed to reach the maximum value of $10$ Mbps, but the encoded data cannot overcome approximately $7.51$ Mbps, given the considered sensor and environment. Hence, the large difference between the target bitrate and throughput when the target bitrate is capped to the maximum value, e.g, $0-100$ s, or $540-860$ s, is accounted to the fact that the encoded data does not reach higher bitrates even in the \textbf{least} encoded state, given (\texttt{Draco}) our algorithm of choice. Specifically, minimum encoding corresponds to $\mathbf{c}_\rho=(24, 0, N)$ in our case, where $N = 32768 \times 3$. Furthermore, based on the results, one may argue that an additional control module such as a simple proportional gain controller can be used to achieve better tracking between the enc. bitrate and $r_\text{trg}$; however the current performance is satisfactory and hence this is regarded to be beyond the scope of this work and thus considered for future work.  
%%%%%%%%%%%%%%%%%%%%%%%%%%%%%%%%%%%%%%%%%%%%%%%%%%%%%%%%%%%%%%%%%%%%%%%%%%%%%%%%
%%%%%%%%%%%%%%%%%%%%%%%%%%%%%%%%%%%%%%%%%%%%%%%%%%%%%%%%%%%%%%%%%%%%%%%%%%%%%%%%
%%%%%%%%%%%%%%%%%%%%%%%%%%%%%%%%%%%%%%%%%%%%%%%%%%%%%%%%%%%%%%%%%%%%%%%%%%%%%%%%
%%%%%%%%%%%%%%%%%%%%%%%%%%%%%%%%%%%%%%%%%%%%%%%%%%%%%%%%%%%%%%%%%%%%%%%%%%%%%%%%
Another essential point relates to the fact that the training of the CPP module manifested from data collected in a completely different environment compared to the evaluation environment. To be precise, the training data is a set of approximately $10$ minutes of LiDAR data, randomly compressed (sampled from a uniform distribution) across the compression parameters described in \eqref{eq:q-param-draco} and \eqref{eq:c-param-draco}, and for stable $N$. The training LiDAR data was recorded in a confined, underground, tunnel environment, while the full algorithm was evaluated in an outdoor open-space public road environment with live vehicle traffic.

The performance of the framework can be also evaluated over the challenging factors that public cellular networks present. Handovers, and the potential varying signal conditions, along with the different cell load experienced when moving from one cell to another are some notable examples. Figure \ref{fig:res:handovers} shows different serving cells that the 5G-enabled robotic application connects to, while Figure \ref{fig:res:radio-kpis} shows the RSRP (Reference Signal Received Power), RSRQ (Reference Signal Received Quality), and RSSI (Received Signal Strength Indicator) over time. Note that we do not have access on the actual cell load on the public network, but we can clearly see the framework adapting to the collective transmission path capacity variations. 

Nevertheless, throughout this particular experiment, the platform experiences various signal conditions, spanning from very good (e.g., RSRP $\geq - 80$ dBm) to quite poor, (e.g., RSRP $\leq - 100$ dBm). However, there is no one-to-one corelation to the end-to-end performance of communication KPIs, nor with robotics KPIs which will be discussed next. This correlation absence is a positive indicator when it comes to the robustness of the method and the fact that it correctly adapts given the sensed transmission path capacity. 

%%%%%%%%%%%%%%%%%%%%%%%%%%%%%%%%%%%%%%%%%%%%%%%%%%%%%%%%%%%%%%%%%%%%%%%%%%%%%%%%
%%%%%%%%%%%%%%%%%%%%%%%%%%%%%%%%%%%%%%%%%%%%%%%%%%%%%%%%%%%%%%%%%%%%%%%%%%%%%%%%

%%%%%%%%%%%%%%%%%%%%%%%%%%%%%%%%%%%%%%%%%%%%%%%%%%%%%%%%%%%%%%%%%%%%%%%%%%%%%%%%
\subsection{Use-case evaluation}
\begin{figure}[!t]
    \includegraphics[width=\linewidth]{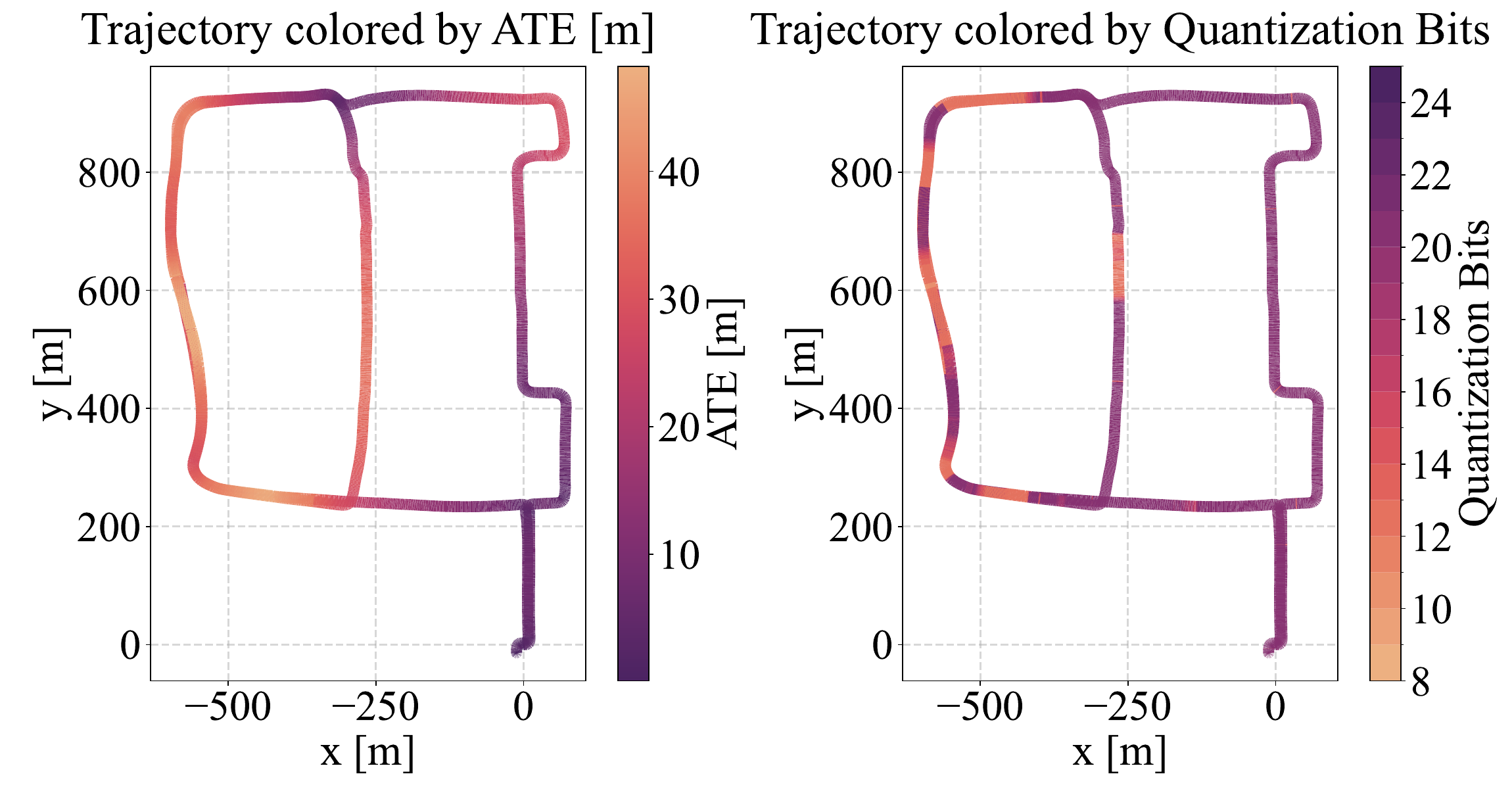}
    \caption{Leftmost subfigure depicts the generated trajectory from the edge offloaded \texttt{KISS-SLAM} algorithm, while the color bar depicts the ATE with ground truth. The right subfigure colorbar shows the corresponding $q$ compression parameter.} 
    \label{fig:res:trajectories}
\end{figure}

%%%%%%%%%%%%%%%%%%%%%%%%%%%%%%%%%%%%%%%%%%%%%%%%%%%%%%%%%%%%%%%%%%%%%%%%%%%%%%%%
Figure \ref{fig:res:trajectories} shows the edge offloaded \texttt{KISS SLAM} by~\cite{guadagnino2025kiss}, produced by all the framework transmitted point clouds in real-time. We present three different metrics. First, to acquire the ground truth trajectory we execute \texttt{KISS SLAM} fully onboard with the uncompressed LiDAR data. The leftmost subfigure depicts the produced trajectory from the offloaded algorithm against the groundtruth where the colorbar indicates the ATE. Here the framework adapts the bitrate in the preconfigured span where the minimum bitrate corresponds to a higher allowed maximum PtP error of $5$ cm, according to \eqref{eq:compression-error-limit}. Therefore, to honor this, the $r_\text{trg}$ span is defined by  $r_{\text{min}}, r_{\text{max}} = 3, 10$ [Mbps]. This value has been shown to produce an RMSE ATE error of $5$ m while capable of achieving larger error values instantaneously (RMSE ATE is shown in Figure \ref{fig:prel:errors}).  

The effect of the $r_\text{trg}$ span is also indirectly visible in the rightmost Figure of \ref{fig:res:trajectories}. The colormap indicates the utilized quantization bits $q$, that consitutes the dominant quantization parameter of $\mathbf{c}_r$. As desired the lowest value of $q = 12$ is not violated. Overall, the required target bitrate is sufficiently tracked by the full framework, and the channel capacity is respected, resulting in the required low-latency transmission while also respecting the posed residual constraints.  

However, such errors can accumulate over time, leading to the higher values captured in the colormap. This behavior can be attributed to several factors. First, SLAM algorithms are inherently stochastic and often require additional techniques to achieve consistent accuracy, such as fusing data from an IMU (Inertial Measurement Unit) to account for agile motion. Another contributing factor is the rapid switching between quantization levels throughout the mission. Although necessary to track the required bitrate, this can pose challenges, as consecutive scans may vary in point density and affect the necessary scan matching procedure in SLAM. However, this issue can be effectively addressed through more advanced encoding algorithms or by accounting for rapid compression changes, as proposed in \cite{Cao-real-time-icra}, while keeping the underlying framework unchanged.

Nevertheless, we chose to evaluate on this simplified version of SLAM, as it is widely recognized within the community and can highlight communication-related challenges that could otherwise be obscured, for instance, when considering IMU fusion and more complex methods.
%%%%%%%%%%%%%%%%%%%%%%%%%%%%%%%%%%%%%%%%%%%%%%%%%%%%%%%%%%%%%%%%%%%%%%%%%%%%%%%%
%%%%%%%%%%%%%%%%%%%%%%%%%%%%%%%%%%%%%%%%%%%%%%%%%%%%%%%%%%%%%%%%%%%%%%%%%%%%%%%%
Lastly, consider the case where the $r_\text{trg}$ feedback is missing, preventing the adjustment of the bitrate required for the queue to drain efficiently. In this situation, compression is controlled by a fixed quantization parameter $q = 16$ (chosen as a moderate value), resulting in a relatively high, non-adaptive bitrate. At times, this bitrate exceeded the channel capacity, leading to a series of packet drops caused by queue overflow. These effects can also be observed in the accompanying video, where the impact of consecutive dropped scans is visible. Such behavior makes the offloading of SLAM or the use of 3D LiDAR data in more time-critical applications, such as collision avoidance, particularly challenging.

%%%%%%%%%%%%%%%%%%%%%%%%%%%%%%%%%%%%%%%%%%%%%%%%%%%%%%%%%%%%%%%%%%%%%%%%%%%%%%%%
%%%%%%%%%%%%%%%%%%%%%%%%%%%%%%%%%% Conclusion %%%%%%%%%%%%%%%%%%%%%%%%%%%%%%%%%%
\section{Conclusions and Future Developments}
\label{conclusions}
%%%%%%%%%%%%%%%%%%%%%%%%%%%%%%%%%%%%%%%%%%%%%%%%%%%%%%%%%%%%%%%%%%%%%%%%%%%%%%%%
In this work, we have successfully demonstrated rate-adaptive streaming of 3D LiDAR data. Since the proposed method builds upon \texttt{SCReAM v2}, which has a strong foundation in rate-adaptation algorithms and literature, the developed approach inherits all the necessary components to achieve low-latency and low-loss performance. To realize this, the encoder module was successfully integrated to ensure that the target bitrate accurately reflected the available channel capacity. In addition, a modeling approach was developed to link the transmission framework’s performance and the minimum allowable transmission bitrate with compression residuals, which capture distance and connect to robotics KPIs.

Finally, it is important to note that even robust rate-adaptive frameworks will underperform when the channel capacity falls below the minimum allowable transmission bitrate, as network, and particularly radio, resources are finite and can be limited in high-load conditions. In such scenarios, leveraging on-demand 5G/6G Quality of Service mechanisms can be critical, as they can help secure the necessary resources and increase the available capacity for the user. These extensions, among others, constitute the main directions for our future work. 

% \begin{itemize}
%     \item \textcolor{orange}{Robustness in large scale, real-world missions and very good fit of the proposed model and our additions.}
%     \item \textcolor{orange}{real-time behavior and robotics performance. Can be used for offloading SLAM, object detection, etc.}
%     \item \textcolor{orange}{Good performance when it comes to the $L_2$ connection to the minimum configured bitrate, etc.}
%     \item \textcolor{orange}{QoD integration, and potentially varying minimum bitrate selection based on mission risk, etc. probably to much to fit}
% \end{itemize}

%%%%%%%%%%%%%%%%%%%%%%%%%%%%%%%%%%%%%%%%%%%%%%%%%%%%%%%%%%%%%%
\bibliography{ifacconf}

%%%%%%%%%%%%%%%%%%%%%%%%%%%%%%%%%%%%%%%%%%%%%%%%%%%%%%%%%%%%%%
\end{document}